\documentclass{article}

\usepackage{graphicx}
\usepackage{amssymb}
\usepackage{amsmath}
\usepackage{amsfonts}
\usepackage{listings}
\usepackage{algorithm}
\usepackage{algpseudocode}
\usepackage{enumitem}
\usepackage{verbatim}
\usepackage{amsthm}
\usepackage{array}
\usepackage[utf8]{inputenc}
\usepackage{caption}
\usepackage{subcaption}
\usepackage{color}

\DeclareMathOperator*{\argmax}{arg\,max}
\DeclareMathOperator*{\argmin}{arg\,min}

\newcommand{\hhf}{{\scriptstyle{{\frac{1}{2}}}}}

\newcommand{\RR}{\mathbb{R}}

\definecolor{myblue}{rgb}{0, 0.2, 0.8}

\newcommand\revise[1]{#1}

\begin{document}

\title{Adversarial Ink: Componentwise Backward Error Attacks on Deep Learning}

\author{Lucas Beerens\thanks{School of Mathematics and The Maxwell Institute for Mathematical Sciences, University of Edinburgh, EH9 3F, UK} \and
Desmond J. Higham\thanks{School of Mathematics, University of Edinburgh, James Clerk Maxwell Building, Edinburgh, UK} } 

\date{}

\maketitle

\abstract{
Deep neural networks are capable of state-of-the-art performance in many classification tasks.
However, they are known to be vulnerable to adversarial attacks---small perturbations 
to the input that lead to a change in classification.
We address this issue
from the perspective of
backward error and condition 
number, concepts that
have proved useful 
 in numerical analysis. To do this, we build 
on the work of Beuzeville et al.\ (2021).
In particular, we develop a new class  
 of attack algorithms
 that use componentwise
relative perturbations. Such attacks are highly relevant in  the case of 
handwritten documents or printed texts where, for example, the classification of 
signatures, postcodes, dates or numerical quantities may be altered by changing only the 
ink consistency and not the background.
This makes the perturbed images look natural to 
the naked eye. 
Such ``adversarial ink'' attacks therefore reveal a weakness that can have a serious 
impact on safety and security. 
We illustrate the new attacks on real data and contrast them with existing 
algorithms. We also study the use of a componentwise condition number  to 
quantify vulnerability.
}


\maketitle

\section{Motivation}
Over the past decade it has become clear that state of the art deep learning image classification tools 
are susceptible to adversarial attacks---deliberately constructed perturbations that are intended to go unnoticed by 
humans but cause a change in the predicted class  \cite{harness,szegedy2013intriguing}.
This type of vulnerability is of concern in high stakes application areas, including 
medical imaging, transport, defence and finance \cite{M18}.
Consequently there has been a great deal of interest in the design of practical attack and defence strategies
\cite{Attack_survey_2018,robust,mmstv18,deepfool,bb}
and, more recently, in theoretical questions concerning the existence and computability   
of adversarial perturbations
\cite{BHV21,fawzi18,shafahi2018adv,tyukin2020adversarial,thwg21}.
 
From the perspective of applied and computational mathematics, the fundamental question to be addressed here concerns well-posedness, or \emph{conditioning}. In particular,  \emph{backward error} theory 
from numerical analysis is highly pertinent.
In \cite{beuzeville:hal-03296180}, the authors used the concept of backward error to construct 
a new form of adversarial attack algorithm.
In this work, we build on this idea by focusing on a special class of data perturbation.
We develop attack strategies based on \emph{componentwise relative} perturbations; for example, each pixel may be perturbed by a small percentage of its original value. 
In particular, this approach allows us to preserve the background of a document and perturb only the 
ink levels in the text.
We also test the corresponding condition number as an indicator of vulnerability to attack.
  
\begin{figure}
    \centering
    \begin{subfigure}[t]{0.3\textwidth}
        \centering
        \includegraphics[width=\textwidth]{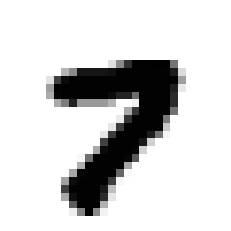}
        \caption{Image from MNIST dataset
        \cite{lcb-digits}, which is correctly classified as a 7 by a neural network.}
    \end{subfigure}
     \hfill
    \begin{subfigure}[t]{0.3\textwidth}
        \centering
        \includegraphics[width=\textwidth]{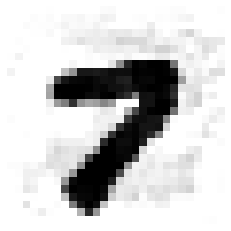}
        \caption{Perturbed image using DeepFool \cite{deepfool}. The background is now smeared. This image is
        classified as an 8.}
    \end{subfigure}
    \hfill
    \begin{subfigure}[t]{0.3\textwidth}
        \centering
        \includegraphics[width=\textwidth]{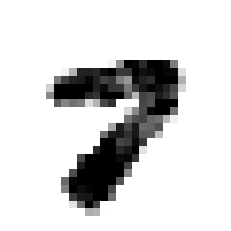}
        \caption{Perturbed image using Algorithm~4 from 
        section~\ref{sec:comp}. The background is unchanged. 
        This image is classified as an 8.}
    \end{subfigure}
    \caption{Image from MNIST and two adversarial attacks.}
    \label{fig:showAttack}
\end{figure}

  To illustrate the main idea, Figure~\ref{fig:showAttack} part (a) 
  shows a handwritten digit from the MNIST data set \cite{lcb-digits}, and 
  parts (b) and (c) show 
  adversarial attacks on a trained network described in section~\ref{sec:comp}.
  The original image (a) is correctly classified as a 7 by the network.
  The perturbed images in (b) are (c) are both classified as an 8.
  For (b) we attacked with the DeepFool algorithm \cite{deepfool}, which 
  controls the Euclidean norm of the perturbation.
  Although the image still has the appearance of a 7,
  we see that 
  the 
  background,
  where pixels had a value of zero,
  is 
  noticeably altered.
  For (c) we computed a componentwise perturbation
  with the new Algorithm~4 from section~\ref{sec:comp} 
  and we see that 
  the background is unchanged.
  The componentwise perturbation in part (c) is
  compatible with a blotchy pen, 
  imperfect paper, or inconsistent handwriting pressure.
  \revise{Indeed, variations in 
  line continuity, line quality and pen control have been widely observed, and are listed among the 
   twenty-one discriminating elements of handwriting \cite{HBS09,HH99}.}
   Hence we argue that this type of ``adversarial ink'' attack produces a more natural result than the 
  perturbation in part (b). 

The manuscript is organized as follows.
In section~\ref{sec:bg} we set up some notation and introduce the 
concept of backward error.
Section~\ref{sec:normwise}
describes the adversarial attack algorithms in 
\cite{beuzeville:hal-03296180}, and Section~\ref{sec:comp}
extends these to the componentwise setting.
Computational results on the MNIST data set 
are presented in Section~\ref{sec:compres}; 
we show illustrative images, summarize the perturbation sizes,
consider untargeted attacks,
compare against state-of-the-art algorithms,
summarize the most likely class changes,
look at the use of a  
condition number to indicate vulnerability to attack,
test on different neural networks, 
and 
report on black box attacks.
Conclusions are given in Section~\ref{sec:conc}.

\section{Image Classification and Backward Error}\label{sec:bg}
We begin by considering a general image classifier, in the form of a map
$ F:[0,1]^n \rightarrow \RR^c$.
Hence, 
we regard an image as a single vector in $\RR^{n}$.
The $n$ components may correspond to individual pixel
intensities in the greyscale case, or red, green, and blue 
channel pixel
intensities in the colour case.
Intensities are assumed to lie in $[0,1]$.
Each image is assigned to one of  $c$ classes, according to 
the largest component of $F(x)$.
In practice, the output vector 
$y = F(x) \in \RR^{c}$ may be passed through a softmax function, so that 
\[
\frac{ e^{y_i} } { \sum_{j=1}^{c} e^{y_j} }
\]
is viewed as the probability that $x$ belongs to class $i$. 
Since $x$ is assigned  to the most likely class,  
we  do not need to include this 
final layer when considering the classification results.

In numerical analysis, 
the concept of backward error deals with the following question:
given an approximate solution, what is the size of the smallest 
perturbation to the input which makes this solution exact?
\revise{
In more detail, suppose an approximation algorithm produces a function ${\widehat H}$, 
instead of the desired function $H$.
For an input $x$, when the algorithm returns ${\widehat H} (x) = y + \Delta y$ instead of $y = H(x)$, we may ask 
for the 
smallest 
$\Delta x$ such that $H  (x  + \Delta x) = y + \Delta y$.
In many settings, the size of $\Delta x$  (the backward error) is 
more  
relevant and more amenable to analysis 
than the size  of $\Delta y$ (the forward error) \cite{CorlessNumerical,high:ASNA2}. 
For an adversarial attack on a classifier, we may interpret $\Delta y$ as a \emph{desired} change in the output.
Then a question of the same structure arises---what is the smallest 
perturbation to the input that achieves the desired output? In this setting, we require  
$\Delta x$ such that $F(x  + \Delta x) = y + \Delta y$.}

This approach was exploited in 
 \cite{beuzeville:hal-03296180}, leading to 
 what we describe as Algorithms~1 and 2 
 in 
 subsections~\ref{subsec:lin} and \ref{subsec:it}.

\section{Normwise Backward Error Attacks: Algorithms~1 
and 2}\label{sec:normwise}
\subsection{Set-up}

In the next two subsections, we describe the 
data perturbation approach from \cite{beuzeville:hal-03296180};
leading to Algorithms~1 and 2.
We cover this existing work in sufficient detail that
(a) there are well-defined algorithms that can be implemented in practice, and
(b)
the
new versions, Algorithms~3 and 4  
in section~\ref{sec:comp}, can be
introduced naturally and compared computationally.

We formulate all algorithms in terms of 
linearly-constrained linear least-squares problems,
for which high quality software is available.
Letting $\| \cdot \|_2$ denote the Euclidean norm, 
these problems have the form
\begin{equation}
    \min_z  \Vert C_1z-k_1\Vert_2,\quad\text{such that}\quad \begin{cases}
        C_2 z\leq k_2,\\
        C_3 z = k_3,
    \end{cases}
    \label{Eq: lsqlin formulation}
\end{equation}
where the matrices $C_1,C_2,C_3$ and vectors
$z,k_1,k_2,k_3$ have appropriate dimensions and where 
vector equalities and inequalities are to be interpreted 
in a componentwise sense.
(More traditionally, the objective function 
in (\ref{Eq: lsqlin formulation}) may be written 
$\hhf \Vert C_1z-k_1\Vert^2_2$, but, of course,
the factor $\hhf$ and the square may be ignored.)
We also assume for now that the Jacobian of the classification map is available; in subsection~\ref{subsec:blackbox} we
test the use of a finite difference approximation
to the Jacobian.

\subsection{Linearized Algorithm}\label{subsec:lin}
We begin by measuring perturbations in the Euclidean norm. 
Given an image $x$ with $F(x) = y$ and a desired new output $\widehat{y}$, a suitable perturbation may be expressed as
\begin{equation}
    \argmin_{\Delta x}\{   \| \Delta x \|_2  \,  :  \,  F(x+\Delta x) = \widehat{y} \}.
    \label{eq:argmin1}
\end{equation}
In general, this problem  cannot be solved analytically.
On the grounds that we are looking for a small perturbation, it is reasonable to linearize, using
 $F(x+\Delta x) - F(x) \approx \mathcal{A} \Delta x$, where 
 $\mathcal{A} \in \RR^{c \times n}$ is the Jacobian of $F$ at $x$, and $F$ is assumed to be 
 differentiable in a neighbourhood of $x$.
 The problem (\ref{eq:argmin1}) then reduces to  
\begin{equation}
    \argmin_{\Delta x}\left\{ \| \Delta x \|_2  \, :  \, \mathcal{A}\Delta x = \widehat{y}-y\right\}.
    \label{eq:linopt}
\end{equation}
For any fixed $ \widehat{y}$  
this is a minimum Euclidean norm linear system.
Typically $c \ll n$, so the system is underdetermined.
Generically, 
a solution for this fixed $\widehat{y}$ can be found by introducing the \revise{ 
Moore-Penrose inverse \cite{generalizedInverse},  
$
    \mathcal{A}^\dagger$}, 
    to give  
\begin{equation}
    \argmin_{\Delta x}\left\{\Vert \Delta x\Vert_2 : \mathcal{A}\Delta x = \widehat{y}-y\right\} = \mathcal{A}^\dagger (\widehat{y}-y).
    \label{eq:Dxsol}
\end{equation}

Given the solution (\ref{eq:Dxsol}), we can use $\widehat{y}$ as an optimization variable.
In the targeted case, where we wish the perturbed image to be 
classified with label  $c_0$, 
we introduce the misclassification set 
\begin{equation}
    \mathcal{S} := \{\widehat{y}\in\mathbb{R}^c \, :  \,  
      \widehat{y}_{c_0}  = \max_{1 \le i  \le c} \widehat{y}_i
      \}.
  \label{eq:Sdef}
\end{equation}
 To compute an adversarial attack we then solve 
\begin{equation}
    \argmin_{\hat{y}\in\mathcal{S}} \Vert \mathcal{A}^\dagger(\hat{y}-y) \Vert_2,
\label{eq:alg1opt}
\end{equation}
and set 
$\Delta x = \mathcal{A}^\dagger (\widehat{y}-y)$.

We now show that 
the problem (\ref{eq:Sdef})--(\ref{eq:alg1opt})
has the form (\ref{Eq: lsqlin formulation}).  
The variable to be optimized is $\widehat{y}$, so we will use $z=\hat{y}$ in \eqref{Eq: lsqlin formulation}.
We may take $C_1 = C_1^{[1]}:= \mathcal{A}^\dagger$ and $k_1 = k_1^{[1]} := \mathcal{A}^\dagger y$. The misclassification condition is equivalent to
    \begin{equation*}
        \hat{y} - \begin{pmatrix}\widehat{y}_{c_0}\\\vdots\\\widehat{y}_{c_0}\end{pmatrix}\leq 0.
    \end{equation*}
    In order to write this in matrix form, we define the matrix $G\in\RR^{c\times c}$ with 
    \begin{equation}
        G_{ij} = \begin{cases}
            1 & \text{ if } j = c_0,\\
            0 & \text{otherwise}.
        \end{cases}
        \label{eq:Gdef}
    \end{equation}
    The misclassification condition may then be written 
    \begin{equation*}
        (I-G)\hat{y} \leq 0.
    \end{equation*}
    Hence we use $C_2 = C_2^{[1]} := I-G$ and 
    $k_2 = k_2^{[1]} = 0$. There are no equality conditions, so we set $C_3 = C_3^{[1]} := 0$ and $k_3 = k_3^{[1]} := 0 $.

This leads to Algorithm~1, as summarized in the displayed pseudocode
listing.

\begin{algorithm}
\caption{Basic normwise attack, returning a perturbed image}\label{alg1.0}
\begin{algorithmic}[1]
\Procedure{Attack}{$F,x,c_0$}
    \State $out \gets F(x)$
    \State $jac \gets$ Jacobian$(F, x)$\label{alg1.0 jac}
    \State $pi \gets$ PseudoInverse$(jac)$
    \State $con \gets$ Constraints$(c_0)$\label{alg1.0 con}
    \State $obj \gets$ Objective$(pi, out)$\label{alg1.0 obj}
    \State $z\gets$ OptimizationVariable
    \State $prob\gets$ Minimize$(z, obj, con)$
    \State Solve$(prob)$
    \State $\Delta y \gets z.value - out$
    \State $\Delta x \gets pi \cdot \Delta y$
    \State \textbf{return} Scale$(F,x,\Delta x,c_0)$ \label{alg1.0 scale}
\EndProcedure
\end{algorithmic}
\end{algorithm} 

The final step of Algorithm~1 requires further explanation.
We note that the constrained linear least-squares problem is not guaranteed to produce a perturbed image 
with values in $[0,1]$. For this reason, we prune the entries using 
\begin{equation*}
    \operatorname{Prune}_i(x) = \begin{cases}
        0 & \text{ if } x_i <0\\
        x_i & \text{ if } x_i\in[0,1]\\
        1 & \text{ if } x_i >1.
    \end{cases}
\end{equation*}
We also note that the resulting $\Delta x$
might be unsuccessful; that is,
$F(x + \Delta x)$ might not
correspond to class $c_0$. 
Hence, we
regard $\Delta x$ 
from 
(\ref{eq:alg1opt})
as a \emph{direction} in which to perturb, and take the smallest
increment that results in class $c_0$. So the Scale function in Algorithm~\ref{alg1.0} is defined as 
\begin{equation}
    \operatorname{Scale}(F,x,\Delta x,c_0) := x + \min\{ a\in\mathbb{R}: \argmax_i F(\operatorname{Prune}(x+a \, \Delta x))_i = c_0\}\, \Delta x.
    \label{eq:Scaledef}
\end{equation}
\revise{
This minimisation is carried out by computing $\argmax_i F(\operatorname{Prune}(x+a \, \Delta x))_i$ over a finely spaced range of $a$ values in 
$(0, \| x \|_2/\| \Delta x\|_2]$. 
If there is no 
suitable value of $a$ in this range 
 then we terminate and regard the attack as unsuccessful. We use this range because for larger $a$ the norm of the perturbation before pruning 
 exceeds the norm of the image, at which point it is reasonable to assume that the perturbation is too large to be of interest.
}

\subsection{Iterative Algorithm}\label{subsec:it}
An alternative approach was also proposed in
\cite{beuzeville:hal-03296180}. This may be motivated by two ideas.
\begin{itemize}
\item ~Do not exploit the analytical solution 
of the linearized problem (\ref{eq:Dxsol}), 
and proceed directly with 
numerical optimization.
This allows us to build in constraints 
that keep all pixel values in the range $[0,1]$.
\item ~Given that we have linearized the problem, take small steps and  proceed iteratively.
\end{itemize}

The algorithm uses a hyperparameter $\alpha$ indicating the step size for each iteration. Again we 
will target some class $c_0$. We start with the perturbation $\Delta x = 0$ and update it with a small multiple of some new $\delta x$ in every step. \revise{At every step we also have $x_\text{new} = x + \Delta x$ and $y_\text{new} = F(x_\text{new})$.} In each of these steps we solve
\begin{equation}
    \argmin_{\delta x} \| \Delta x + \delta x\|_2,
    \label{eq:dxstep}
\end{equation}
under the misclassification condition
\revise{$\widehat{y} \in  \mathcal{S}$,
the condition that $\mathcal{A}\delta x = \widehat{y} - y_\text{new}$,} and the constraint that pixel values lie in the unit interval. 
Since we now have constraints on both $\widehat{y}$ and $\delta x$, we treat them both as optimization variables.
Then we update $\Delta x$ and \revise{$x_\text{new}$} by adding $\alpha\delta x$ to both. Finally $\mathcal{A}$ is recomputed before moving on to the next iteration.

To see that we still have constrained linear 
least-squares problems of the form 
 (\ref{Eq: lsqlin formulation}), note that we must repeatedly solve
 (\ref{eq:dxstep})
    under the conditions that \revise{$\mathcal{A}\delta x = \widehat{y} - y_\text{new}$} and
      $ \widehat{y} \in  \mathcal{S}$.
     Since both $\delta x$ and $\widehat{y}$ need to be optimized we use 
    $z= \begin{bmatrix}
    \delta x \\ \widehat{y}
    \end{bmatrix}$
    in \eqref{Eq: lsqlin formulation}. 
    Thus we use 
    $C_1 = C_1^{[2]} := \begin{bmatrix}
    I, 0
    \end{bmatrix}$
    and $k_1 = k_1^{[2]} := -\Delta x$. \revise{For the inequality constraints we need to consider the misclassification constraint and the pixel value bound constraint.
    Keeping in mind 
 that $z$ also includes $\delta x$
    we obtain 
    $\begin{bmatrix}
    0, I-G
    \end{bmatrix}z\leq0$, for $G$ in (\ref{eq:Gdef}). }
    Pixel values must also lie in the unit interval. This constraint may be written as
    \begin{equation*}
        \begin{bmatrix}
            I& 0\\
            -I& 0
        \end{bmatrix}z 
        \leq
        \revise{\begin{bmatrix}
            \mathbf{1}-x_\text{new}\\
            x_\text{new}
        \end{bmatrix}},
    \end{equation*}
    where $\mathbf{1}$ denotes a vector of $1$s.
    Combining the two inequality conditions gives 
    \begin{equation*}
        C_2 = C_2^{[2]} := \begin{bmatrix}
            0&I-G\\
            I& 0\\            
            -I& 0
        \end{bmatrix}\quad \text{and}\quad
        k_2 = k_2^{[2]} := \revise{\begin{bmatrix}
            0\\
            \mathbf{1}-x_\text{new}\\
            x_\text{new}
        \end{bmatrix}}.
    \end{equation*}
    The required equality condition is $\mathcal{A}\delta x = \widehat{y} - \revise{y_\text{new}}$; that is,  
    $\begin{bmatrix}
    -\mathcal{A},I
    \end{bmatrix}z = \revise{y_\text{new}}$.
    Therefore we take 
    $C_3 =C_3^{[2]} := \begin{bmatrix}
    -\mathcal{A},I
    \end{bmatrix}$
    and $k_3= k_3^{[2]} := \revise{y_\text{new}}$.

This leads to Algorithm~2, 
summarized in displayed the pseudocode.
Here we have a prescribed number of iterations, \emph{num}. 
In Section~\ref{sec:comp} we examine the performance in terms of 
the iteration number.

\begin{algorithm}
\caption{Iterative normwise attack, returning a perturbed image}\label{alg2.0}
\begin{algorithmic}[1]
\Procedure{Attack}{$F,x,c_0,\alpha,num$}
    \State $out \gets F(x)$
    \State $\Delta x\gets 0$
    \State $newX \gets x$
    \For{$i = 1$ \textbf{to} $num$}
        \State $jac \gets$ Jacobian$(F, newX)$
        \State $pi \gets$ PseudoInverse$(jac)$
        \State $con \gets$ Constraints$(jac,newX,out,c_0)$
        \State $obj \gets$ Objective$(\Delta x)$
        \State $z\gets$ OptimizationVariable
        \State $prob\gets$ Minimize$(z, obj, con)$
        \State Solve$(prob)$
        \State $\delta x \gets z.value.delta$
        \State $\Delta x \gets \Delta x + \alpha \cdot \delta x$
        \State $newX \gets newX + \alpha \cdot \delta x$
        \State $out \gets F(newX)$
    \EndFor
    \State \textbf{return} Scale$(F,x,\Delta x,c_0)$ \label{alg2.0 scale}
\EndProcedure
\end{algorithmic}
\end{algorithm}

\section{Componentwise Backward Error Attacks: Algorithms~3 and 4}\label{sec:comp}
\label{subsec:compa}
\revise{The minimum Euclidean norm perturbation in (\ref{eq:argmin1}) was motivated by a normwise concept of backward error.
Based on the alternative componentwise backward error viewpoint in 
 \cite{higham_backward_1992,high:ASNA2}, instead of (\ref{eq:argmin1}) we may consider the problem 
\begin{equation*}
    \argmin_{\Delta x}\{\epsilon : F(x+\Delta x) = \widehat{y},\ \  | \Delta x|\leq \epsilon  f\},
\end{equation*} 
for a given tolerance vector $f\geq 0 \in \RR^n$.} Here, the absolute value function $| \cdot |$ is applied to each component, so $|\Delta x|_i$ is  $| \Delta x_i|$. 
Unless otherwise indicated, we will use $f=|x|$.
In this case, changes are measured in 
a \emph{relative componentwise} sense, and, in particular, a zero element of $x$ cannot be perturbed. 

Using this type of constraint in an adversarial attack,
after linearizing, a componentwise version 
of (\ref{eq:linopt}) is given by \revise{
\begin{equation}
    \argmin_{\Delta x}\{\epsilon : \mathcal{A}\Delta x = \widehat{y} - y, \quad |\Delta x|\leq \epsilon f \}.
    \label{eq:lcomp2}
\end{equation}
We now write the constraint in a form that fits into the linear optimization framework 
(\ref{Eq: lsqlin formulation}), using an idea from \cite[Section~2]{higham_backward_1992}. We set $\Delta x = Dv$, where $D=\operatorname{diag}(f)$ and $v$ is a vector. The relevant optimization problem is
\begin{equation*}
   \min\{\epsilon : \mathcal{A}Dv = \widehat{y} - y, \quad |Dv|\leq \epsilon f,\quad Dv=\Delta x\}.
\end{equation*}
Since $D=\operatorname{diag}(f)$, we know that the smallest such $\epsilon$ will always be equal to $\Vert v\Vert_\infty$. Hence the minimization problem can be written}
\begin{equation}
    \min\{\Vert v\Vert_\infty : \mathcal{A}Dv = \widehat{y} - y\}.
    \label{eq:lincomp}
\end{equation}
In the absence of an analytical solution to 
(\ref{eq:lincomp}), we will proceed with an iterative 
algorithm, along the lines of Algorithm~2,
using $v$ and $\delta v$ in place of $\Delta x$ and $\delta x$, respectively. \revise{Again we use $x_\text{new}$ and $y_\text{new}$ to keep track of the perturbed image and output during the iterative process.} In each iteration of the algorithm we compute
\begin{equation*}
    \argmin_{\delta v} \Vert v + \delta v\Vert_\infty,
\end{equation*}
under the conditions that
$  \widehat{y} - \revise{y_\text{new}} = \mathcal{A}D\delta v $
and 
$  \widehat{y} \in \mathcal{S}$.
After each step we update $v\leftarrow v+\alpha\delta v$, where $\alpha$ is a hyperparameter. Then we assign \revise{$\Delta x = Dv$} and \revise{$x_\text{new} = x+\Delta x$}. Finally we recompute $\mathcal{A}$.

To fit into the least-squares framework 
\eqref{Eq: lsqlin formulation}, we introduce a new variable $u\in\mathbb{R}$. To minimize the infinity norm of $v + \delta v$, we may solve 
\begin{equation*}
    \min\{|u|: |v+\delta v|\leq u\mathbf{1}\}.
\end{equation*}
Here, the constraint may be written as two separate linear inequality constraints.
We must also include $\delta v$ and $\hat{y}$ in the optimization variable. We will write 
\begin{equation*}
    z = \begin{bmatrix}u\\\hat{y}\\\delta v\end{bmatrix}.
\end{equation*}
We can now specify the required matrices in \eqref{Eq: lsqlin formulation}. Since the target function is $|u|$, we use $C_1 = C_1^{[3]} := \begin{bmatrix}
1, 0,\hdots, 0
\end{bmatrix}$
and $k_1 = k_1^{[3]} := 0$.
There are five inequality constraints. Two are $\delta v - u  \mathbf{1} \leq -v$ and $-\delta v-u \mathbf{1} \leq v$ coming from the infinity norm optimization. 
A third inequality constraint is 
$  \widehat{y} \in \mathcal{S}$, 
 which we may write as $(I-G)\hat{y}\leq 0$. 
Finally, to keep the pixel values of the perturbed image within the unit interval we require 
\begin{equation*}
    \begin{bmatrix}
            0& \alpha D\\
            0& -\alpha D
        \end{bmatrix}z 
        \leq
        \revise{\begin{bmatrix}
            \mathbf{1}-x_\text{new}\\
            x_\text{new}
        \end{bmatrix}.}
\end{equation*}
Combining these five constraints we obtain
\begin{equation*}
    C_2 = C_2^{[3]} := \begin{bmatrix}
        -\mathbf{1}&0&I\\
        -\mathbf{1}&0&-I\\
        0&I-G&0\\
        0&0& \alpha D\\
        0&0& -\alpha D
    \end{bmatrix},
\end{equation*}
and
\begin{equation*}
    k_2 = k_2^{[3]} :=\revise{ \begin{bmatrix}
        -v\\v\\0\\\mathbf{1}-x_\text{new}\\x_\text{new}
    \end{bmatrix}.}
\end{equation*}
There is also an equality constraint given by 
$\widehat{y} -\mathcal{A}D\delta v  = \revise{y_\text{new}}$. This results in $C_3 = C_3^{[3]} :=  \begin{bmatrix}0&I&-\mathcal{A}D\end{bmatrix}$ and 
$k_3 = k_3^{[3]} := \revise{y_\text{new}}$.

This leads to Algorithm~3, 
which is summarized in the displayed pseudocode.

\begin{algorithm}
\caption{Iterative componentwise attack, returning a perturbed image}\label{alg3.0}
\begin{algorithmic}[1]
\Procedure{Attack}{$F,x,c_0,\alpha,num,f$}
    \State $out \gets F(x)$
    \State $v\gets 0$
    \State $\Delta x\gets 0$
    \State $newX \gets x$
    \State $D \gets Diagonal(f)$
    \For{$i = 1$ \textbf{to} $num$}
        \State $jac \gets$ Jacobian$(F, newX)$
        \State $pi \gets$ PseudoInverse$(jac)$
        \State $con \gets$ Constraints$(jac,v,out,D,c_0)$
        \State $obj \gets$ Objective$(\Delta x)$
        \State $z\gets$ OptimizationVariable
        \State $prob\gets$ Minimize$(z, obj, con)$
        \State Solve$(prob)$
        \State $\delta v \gets z.value.dv$
        \State $v \gets v + \alpha \cdot \delta v$
        \State $\Delta x \gets Dv$
        \State $newX \gets x + \alpha \cdot \Delta x$
        \State $out \gets F(newX)$
    \EndFor
    \State \textbf{return} Scale$(F,x,\Delta x,c_0)$
\EndProcedure
\end{algorithmic}
\end{algorithm}

One issue with Algorithm~\ref{alg3.0} is that 
the problem (\ref{eq:lincomp}) encourages all components of $v$ to achieve the maximum 
$\| v \|_\infty$.
As we will see in section~\ref{sec:comp}, 
this may lead to 
perturbations that are very noticeable.
We therefore consider an alternative version where (\ref{eq:lincomp}) is changed to
\begin{equation}
    \min\{\Vert Dv\Vert_2 : \mathcal{A}Dv = \widehat{y} - y\}.
    \label{eq:lincomp2}
\end{equation}
Because $\Delta x=Dv$, we retain the masking effect where zero values in the tolerance 
vector $f$ force the corresponding pixels to 
remain unperturbed. 
We found that minimizing
$\Vert Dv\Vert_2$ rather than $\| v \|_\infty$
produced perturbations that appeared less obvious.
We will refer to this version as Algorithm~4. 
It differs from Algorithm~\ref{alg3.0} only in that
\revise{$C_1^{[3]}$} is changed to
$
\begin{bmatrix}
    0, 0, D
    \end{bmatrix}$
 and 
$k_1^{[3]}$ is changed to $-Dv$.

\section{Computational Results}\label{sec:compres}

We implemented the algorithms in Python using 
PyTorch \cite{PyTorch19} and tested them in a deep learning setting. For the constrained least-squares optimization, we used the Splitting Conic Solver \cite{scsOptimizer} from the CVXPY Python package \cite{agrawal2018rewriting,diamond2016cvxpy}. 
To evaluate the Jacobian of the classification map, 
 we used the PyTorch function torch.autograd.functional.jacobian, 
 which implements an efficient backpropagation process.

 We tested on the MNIST dataset of handwritten digits \cite{lcb-digits}. All images are $28\times28$ pixels in greyscale.
 They have a black background,
 corresponding to a pixel value of zero,
 with white writing. 
 Hence, choosing a tolerance vector of $f=x$ 
 in Algorithms~3 and 4 causes the background to remain unperturbed.
 After applying the algorithms, we display results in the reverse  greyscale, so that the background appears white and the ``ink'' appears grey-black, which we believe is a more realistic view.
 This dataset consists of 60,000 training images and 10,000 testing images.  Following 
 \cite{beuzeville:hal-03296180},
the network that we use has $784$ input nodes and a hidden layer of $100$ nodes followed by the output layer of $10$ nodes. The layers are fully connected and the first one has a $\tanh$ activation function,
chosen because it is differentiable. We consider different activation functions and network architectures in 
subsection~\ref{subsec:arch}. All network training is done using the Adam optimizer and a cross-entropy loss function. The accuracy of the trained network on the test data is $97\%$.

\subsection{Iterations}
\begin{figure}
    \centering
    \includegraphics[width=\textwidth]{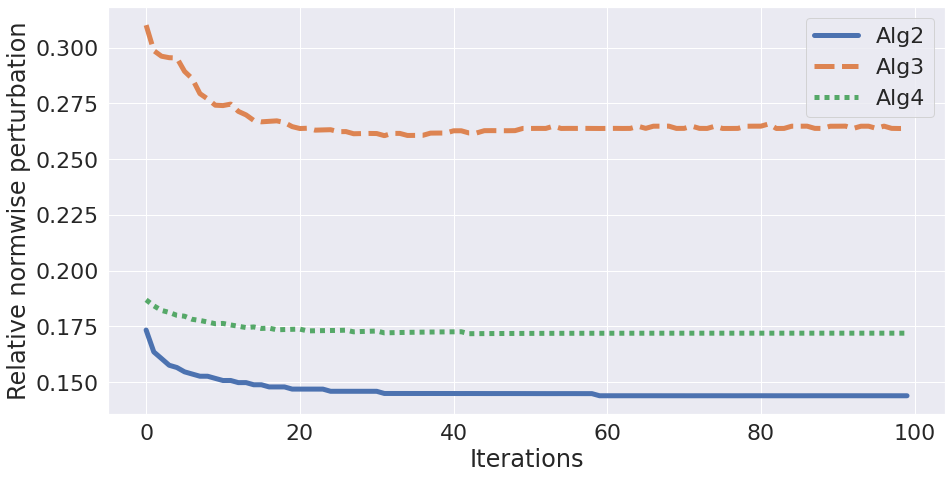}
    \caption{Performance per iteration for Algorithms 2, 3, and 4.
    }
    \label{fig: iterations basic}
\end{figure}

First we investigate how 
Algorithms~2, 3 and 4
perform with respect to the iteration count.
We use $\alpha = 0.1$. After each iteration, using the 
Scale function in 
(\ref{eq:Scaledef}) 
we
take the smallest successful multiple of the direction produced by the algorithm and record the resulting normwise perturbation size, 
$\epsilon = \| \Delta x \|_2 / \| x \|_2$.
In other words, we record 
the $\epsilon$ arising if 
we terminate at that iteration. 
Figure~\ref{fig: iterations basic} 
shows results for a single image.
The horizontal axis gives the iteration count and the vertical axis
gives $\epsilon$.
 The curves, which are similar for other images, indicate 
 that we should use about 30 iterations to get 
 optimal performance; hence we use this value 
 in subsequent experiments.
 We also note that Algorithm~1 behaves in a similar way to the 
 first step of Algorithm 2, and hence 
 Figure~\ref{fig: iterations basic} shows that iterating can give a significant benefit.
Algorithm~3 produces larger relative two-norm perturbations 
that do not decrease monotonically with respect to the iteration count.
This is to be expected, because the algorithm is optimizing $\Vert v\Vert_\infty$. 

Next we show examples of the three iterative algorithms 
successfully attacking images. 
We chose the first image from each digit class, ``0'',``1'',``2'',\ldots,``9'', arising in the training set and 
systematically targeted each incorrect class.
Full results can be seen in the Supplementary Materials section.
In Figure~\ref{fig:imageComp1.2} we have picked out one example for 
attacked images in classes ``5'' to ``9''. In each case we show the perturbed,
incorrectly classified, images from Algorithms~2, 3 and 4.
We also show the size of $\| \Delta x\|_2/\| x\|_2$.
We see that Algorithm~2 perturbs the background whereas,
by construction, Algorithms~3 and 4 do not. This leads to the background looking dirty or smudgy using Algorithm~2. Whenever Algorithm~3 decides it can perturb the pixels by some relative amount,
due to the use of the infinity-norm it does not matter how many pixels are perturbed by that relative amount. This leads to large areas where the black is turned to grey, which is quite noticeable. 
Algorithm~4 addresses this problem by optimising for the 2-norm, as 
shown in \eqref{eq:lincomp2}.
Overall, we see that Algorithm~4 produces images that 
may have arisen naturally from a slight inconsistency in the 
ink delivery or the pen pressure.


\begin{figure}
    \centering
    \includegraphics[width=0.75\textwidth]{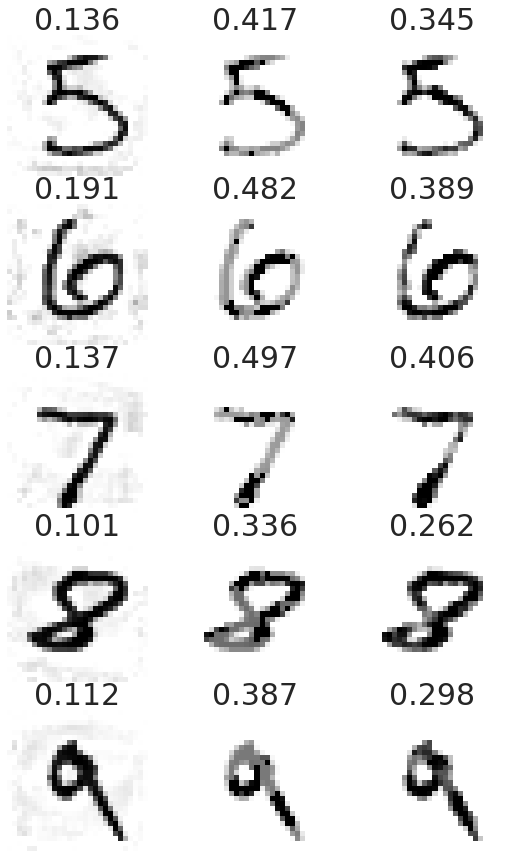}
    \caption{Comparison of successful attacks created using 
    Algorithms~2, 3, and 4, from left to right, with the relative 2-norm of the 
    perturbation indicated.
     Here the original $\rightarrow$ target examples are
     $5 \rightarrow 9$,
     $6 \rightarrow 1$,
     $7 \rightarrow 5$,
     $8 \rightarrow 7$, and
     $9 \rightarrow 3$.
     Examples for digits 0 to 9 and 
     all possible choices of target are shown in the 
     Supplementary Materials  section.
    }
    \label{fig:imageComp1.2}
\end{figure}

\begin{figure}
    \centering
    \includegraphics[width=\textwidth]{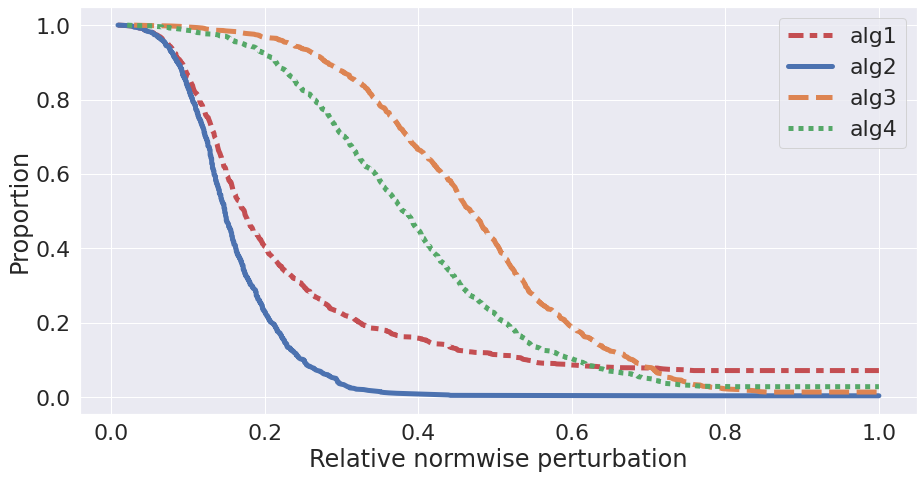}
    \caption{Comparison between relative 2-norm performances of \revise{targeted versions of} Algorithms~1,2,3 and 4. The horizontal axis is the relative 2-norm of the perturbation. The vertical axis is the proportion of attacks requiring at least that relative norm of the perturbation 
     to produce the desired classification.}
    \label{fig:plotComp1}
\end{figure}

In Figure~\ref{fig:plotComp1} we return to the comparison of 
perturbation sizes. 
Here the horizontal axis shows the relative 2-norm of the perturbation. The vertical axis shows the proportion of attacks requiring at least that relative norm of the perturbation to produce the desired classification. 
So a lower curve indicates better performance.
The figure is based on 100 images, resulting in 900 attacks. Algorithm~2 performs best according to this measure. The iterations are seen to significantly improve the 
performance of these targeted attacks---recall that Algorithm~1 does not iterate. Algorithm~3 performs quite poorly, as 
is to be expected, since it does not directly control the 2-norm. 
Algorithm~4, which accounts for the 2-norm while restricting
to componentwise attacks shows better performance in this regard.

\subsection{Untargeted Attacks}\label{subsec:untarg}

\begin{figure}
    \centering
    \includegraphics[width=\textwidth]{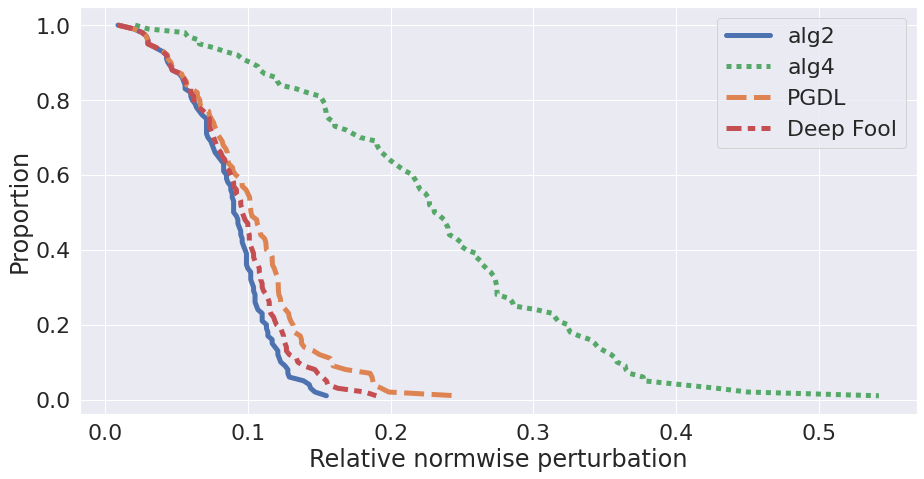}
    \caption{Comparison between the performances \revise{untargeted versions of} of Algorithm 2, Algorithm 4, DeepFool and PGD. 
     Axes as for Figure~\ref{fig:plotComp1}.}
    \label{fig:plotCompareOthers}
\end{figure}

We now consider the scenario where it is sufficient for an attack 
to change the classification to \emph{any} new class.
We deal with this by 
targeting all new classes individually and picking the
smallest perturbation.
We also compare the algorithms with existing approaches designed for this untargeted case.
In Figure~\ref{fig:plotCompareOthers} we compare Algorithms~2 and 
4 with DeepFool \cite{deepfool} and the $\ell_2$ version of 
projected gradient descent (PGD) \cite{mmstv18},
with the performance measure used for Figure~\ref{fig:plotComp1}.
The perturbations 
produced by DeepFool and PGD are scaled in the same manner as that 
described for Algorithms~1, 2, 3, and 4. We see that Algorithm~2
gives the best results. We suggest that this slight improvement 
over Deepfool and PGD arises from (a) the use of $\widehat y$ as
an ``outer'' optimization variable and (b) the use of an iterative procedure to improve the accuracy of the linearizations.
These results confirm that Algorithms~3 and 4 are building on 
a state-of-the-art methodology.

\subsection{Best Targets}\label{subsec:best_targ}

Next, we look at the best targets for each class. To do this we attack 1000 images and save both the original image class 
and the new class of the successful attack with smallest perturbation. 
We then record,
for each original class, 
the proportion of times that each possible new class
arose as the best target. 
Tables~\ref{tab:alg2}  and 
 \ref{tab:alg4} 
show the best target proportions for Algorithms~2 and 4, respectively.
Corresponding results for Algorithms~1 and 3 are shown in the 
Supplementary Materials section:
results for Algorithm~1 are similar to those for 
Algorithm~2, and 
results for Algorithm~3 are similar to those for 
Algorithm~4.
The rows represent the classes of images that we perturb, and the columns represent the target classes. 
So, for example, in Table~\ref{tab:alg2} we see that when images of 
the digit ``0'' were attacked with Algorithm~2, in 38\% of the cases
the smallest perturbation arose 
when the class was changed to ``5.''
The highest proportion for each original class is 
highlighted in bold. 
When comparing the results in Tables~\ref{tab:alg2} and \ref{tab:alg4},
we should keep in mind that Algorithm~2
may take away ink from the digits \textbf{and} add ink to the background, whereas 
Algorithm~4 may only take away ink.
For the digit class ``3'' 
it is notable that Algorithm~2
favours the target class ``5'', with a proportion  
of $0.57$, and the other 
significant target classes are ``2'' and ``8.''
Algorithm~4 is less likely to
perturb from class ``3'' into class ``5''
(the proportion is 0.48) and target class ``7'' has become more frequent (0.10 compared with 0.04 in
Algorithm~2).
It is intuitively reasonable that 
convincingly 
changing a 3 into a 5 or a 2 benefits from both addition and removal 
of ink and  
changing a 3 into a 8 benefits from addition alone, both of which 
are 
natural for Algorithm~2.
Changing 3 into a 7 is 
more of a subtractive process, which suits  
Algorithm~4.
We also see that for class ``1'', Algorithm~2 favours target classes ``3'' and ``7'', which are likely to
require addition of ink, whereas Algorithm~4 has a fairly even spread of proportions---there is no obvious
way to remove ink from a ``1'' in order to approximate a different digit.
Perhaps less obvious are the results for class ``9.''
Here, Algorithm~2 prefers the target class ``7'', with proportion $0.42$, whereas 
Algorithm~4 prefers class ``4'', with proportion $0.63$ and has class ``7'' in second place with proportion $0.24$.
We believe that this effect is explained by the 
fact that there are two widely used versions of the written digit 4.
The version illustrated in the Supplementary Materials section 
is close to the digit 9 with the upper portion of the loop 
removed. 

\begin{table}[h]
    \centering
    \begin{tabular}{c|*{10}c}
    & \multicolumn{10}{c}{Alg.\ 2: best target class proportion}\\
    & 0 & 1 & 2 & 3 & 4 & 5 & 6 & 7 & 8 & 9\\ \hline
    0 & 0.00 & 0.00 & 0.16 & 0.05 & 0.00 & \bf{0.38} & 0.14 & 0.13 & 0.00 & 0.15 \\
    1 & 0.00 & 0.00 & 0.04 & \bf{0.42} & 0.00 & 0.10 & 0.07 & 0.28 & 0.08 & 0.02 \\
    2 & 0.03 & 0.03 & 0.00 & \bf{0.57} & 0.00 & 0.02 & 0.07 & 0.09 & 0.19 & 0.01 \\
    3 & 0.01 & 0.02 & 0.19 & 0.00 & 0.00 & \bf{0.57} & 0.01 & 0.04 & 0.15 & 0.02 \\
    4 & 0.01 & 0.00 & 0.05 & 0.01 & 0.00 & 0.00 & 0.04 & 0.15 & 0.10 & \bf{0.64} \\
    5 & 0.01 & 0.00 & 0.02 & \bf{0.53} & 0.02 & 0.00 & 0.10 & 0.13 & 0.13 & 0.05 \\
    6 & 0.02 & 0.01 & 0.22 & 0.01 & 0.02 & \bf{0.52} & 0.00 & 0.09 & 0.02 & 0.07 \\
    7 & 0.01 & 0.02 & \bf{0.4} & 0.31 & 0.01 & 0.00 & 0.00 & 0.00 & 0.02 & 0.24 \\
    8 & 0.01 & 0.01 & 0.31 & \bf{0.4} & 0.01 & 0.09 & 0.02 & 0.08 & 0.00 & 0.06 \\
    9 & 0.00 & 0.01 & 0.02 & 0.09 & 0.23 & 0.09 & 0.00 & \bf{0.42} & 0.15 & 0.00 \\
    \end{tabular}
    \caption{Table of best target proportion for attacks made by Algorithm 2. The rows are the original classes and the columns are the target classes.}
    \label{tab:alg2}
\end{table}

\begin{table}[h]
    \centering
    \begin{tabular}{c|*{10}c}
    & \multicolumn{10}{c}{Alg.\ 4: best target class proportion}\\
    & 0 & 1 & 2 & 3 & 4 & 5 & 6 & 7 & 8 & 9\\ \hline
    0 & 0.00 & 0.00 & 0.08 & 0.01 & 0.00 & \bf{0.49} & 0.16 & 0.11 & 0.01 & 0.13 \\
    1 & 0.02 & 0.00 & \bf{0.19} & 0.18 & 0.00 & 0.06 & 0.12 & 0.15 & \bf{0.19} & 0.08 \\
    2 & 0.06 & 0.05 & 0.00 & \bf{0.52} & 0.03 & 0.02 & 0.09 & 0.10 & 0.13 & 0.00 \\
    3 & 0.03 & 0.02 & 0.14 & 0.00 & 0.05 & \bf{0.48} & 0.02 & 0.10 & 0.12 & 0.04 \\
    4 & 0.03 & 0.01 & 0.08 & 0.03 & 0.00 & 0.03 & 0.02 & 0.20 & 0.25 & \bf{0.36} \\
    5 & 0.06 & 0.01 & 0.01 & \bf{0.4} & 0.06 & 0.00 & 0.07 & 0.06 & 0.26 & 0.08 \\
    6 & 0.06 & 0.02 & 0.13 & 0.00 & 0.29 & \bf{0.39} & 0.00 & 0.02 & 0.07 & 0.01 \\
    7 & 0.01 & 0.03 & 0.20 & \bf{0.38} & 0.10 & 0.02 & 0.00 & 0.00 & 0.03 & 0.24 \\
    8 & 0.03 & 0.01 & 0.14 & \bf{0.28} & 0.09 & 0.24 & 0.03 & 0.06 & 0.00 & 0.11 \\
    9 & 0.00 & 0.01 & 0.00 & 0.02 & \bf{0.63} & 0.02 & 0.00 & 0.24 & 0.08 & 0.00 \\
    \end{tabular}
    \caption{Table of best target proportion for attacks made by Algorithm 4. The rows are the original classes and the columns are the target classes.}
    \label{tab:alg4}
\end{table}

\subsection{Condition Numbers}\label{subsec:cond_nums}
The backward error concept discussed in Section~\ref{sec:bg}
is traditionally accompanied by a corresponding 
concept of conditioning (or well-posedness). 
A condition number measures the worst-case sensitivity of the output 
to small changes in the input and, by construction, the 
forward error is approximately bounded by the product of a 
condition number and a backward error measure 
\cite{GVl96,higham_backward_1992,high:ASNA2}.
It follows that 
when we use a neural network to classify an image, we may also 
compute an appropriate condition number estimate in order to 
get a feel for the sensitivity of the output to 
worst-case perturbations in the input, and hence to adversarial attacks.
We note, however, that in the experiments reported so far, realistic attacks were very likely to exist for any input, 
and hence we view the condition number as a possible means to 
quantify the \emph{relative} sensitivity.

In the normwise case, if $\epsilon = \| \Delta x \|_2 / \| x \|_2$
is small then 
\[
\frac{
\| F(x) - F (x + \Delta x) \|_2
}
{
\| F(x) \|_2
}
\approx
\frac{
\|
\mathcal{A}  \Delta x \|_2
}
{
\| F(x) \|_2
}
\lessapprox
\frac{
\|
\mathcal{A}
\|_2
\|
\| x \|_2
}
{
\| F(x) \|_2
}
\epsilon
=:
\mu_2(x) \epsilon.
\]
Here, $\mu_2(x)$ may be viewed as a relative normwise condition number. 

Similarly, under the constraint
$|\Delta x | \le \epsilon f$, where we recall that $f$ is a  nonnegative tolerance vector, 
we have 
\[
\frac{
\| F(x) - F (x + \Delta x) \|_\infty
}
{
\| F(x) \|_\infty
}
\approx
\frac{
\|
\mathcal{A}  \Delta x \|_\infty
}
{
\| F(x) \|_\infty
}
\lessapprox
\frac{
\|
|\mathcal{A}| \, f
\|_\infty
}
{
\| F(x) \|_\infty
}
\epsilon
=:
\mu_\infty(x) \epsilon;
\]
so $\mu_\infty(x)$ may be viewed as a relative 
componentwise condition number.

\begin{figure}
    \centering
    \includegraphics[width=0.5\textwidth]{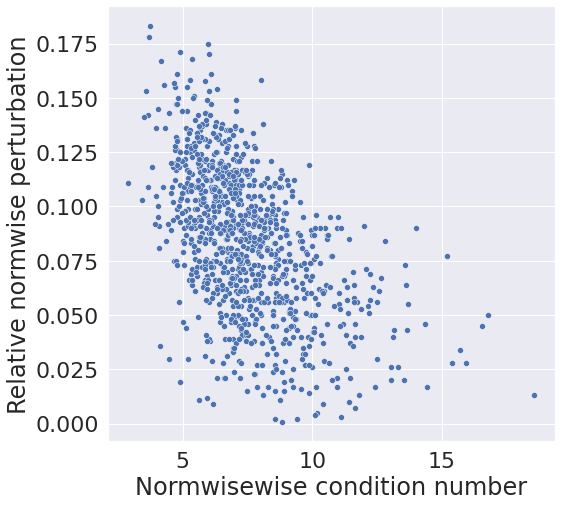}
    \caption{Scatter plot of relative normwise 
    perturbation for Algorithm~2 against normwise condition number $\mu_2$ for 1000 images.}
    \label{fig:condPlotNorm}
\end{figure}

For the normwise case, in Figure~\ref{fig:condPlotNorm}
we use a collection of 1000 test images that are classified correctly.
For each image we compute the best attack from 
Algorithm~2.
The figure scatter plots the attack perturbation size against the 
normwise condition number, $\mu_2$.
We see that a larger value of $\mu_2$ generally
 corresponds,
albeit weakly, to 
a smaller perturbation.
The correlation coefficient is $-0.52$. 

\begin{figure}
    \centering
    \begin{subfigure}{0.49\textwidth}
        \includegraphics[width=\textwidth]{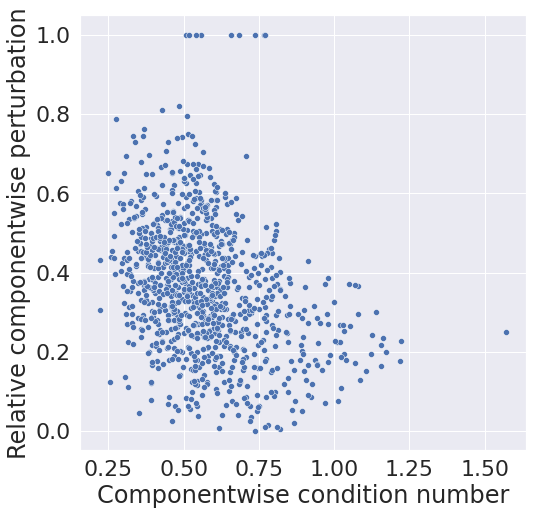}
        \caption{Algorithm 3}
    \end{subfigure}
    \begin{subfigure}{0.49\textwidth}
        \includegraphics[width=\textwidth]{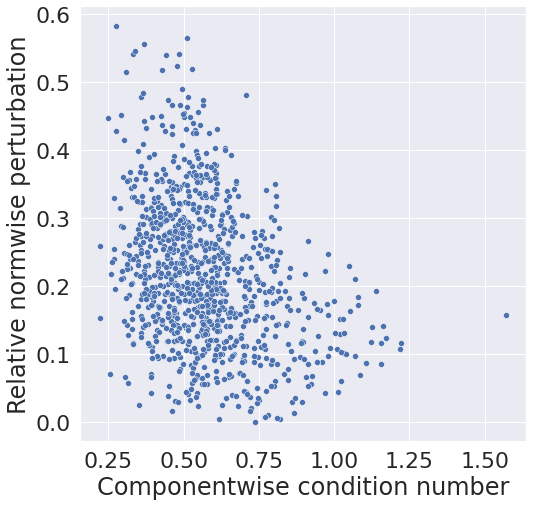}
        \caption{Algorithm 4}
    \end{subfigure}
    \caption{Scatter plot of relative componentwise perturbation and relative 2-norm respectively from 
    Algorithms~3 and 4 against componentwise condition number $\mu_\infty$ for 1000 images.}
    \label{fig:condPlotComp}
\end{figure}

Figure~\ref{fig:condPlotComp} shows corresponding results for 
the componentwise condition number $\mu_\infty$ with attacks from 
Algorithms~3 and 4. We compare this condition number with the performances corresponding to these two attacks: the relative infinity norm and relative 2-norm respectively.
Here, the correlation coefficients are $-0.33$ and $-0.34$ respectively, so the 
condition number is less useful in this case.
A possible explanation for this 
difference is that
perturbations are larger, and hence the 
linearizations are less accurate.

\subsection{Architecture}\label{subsec:arch}
So far we used a two layer network with a $\tanh$ activation function. Let us call this Net1. 
We now consider two further networks.
 Net2 denotes the network arising when 
 $\tanh$ in Net1 is replaced with a rectified linear unit (ReLu). 
We note that ReLu is not differentiable at the origin;
this did not cause any issues in our tests. 
 After training, Net2 has an accuracy of $97$\%. 
 Net3 is a convolutional neural network (CNN) \cite{deepLearningGoodfellow} 
 with two convolutional layers that include ReLu and max pool. The first convolutional layer uses  Conv2d$(1,16,5,1,2)$ in PyTorch. The parameters correspond to the number of input channels, number of output channels, the kernel size, the stride and the padding, respectively. This is followed by ReLu and MaxPool with stride 2. The second layer uses Conv2d$(16,32,5,1,2)$ and is again followed by ReLu and Maxpool with stride 2. The final layer is a fully connected layer leading to an output in $\mathbb{R}^{10}$. 
 Net3 gave an accuracy of $99$\%.

\begin{figure}
    \centering
    \includegraphics[width=\textwidth]{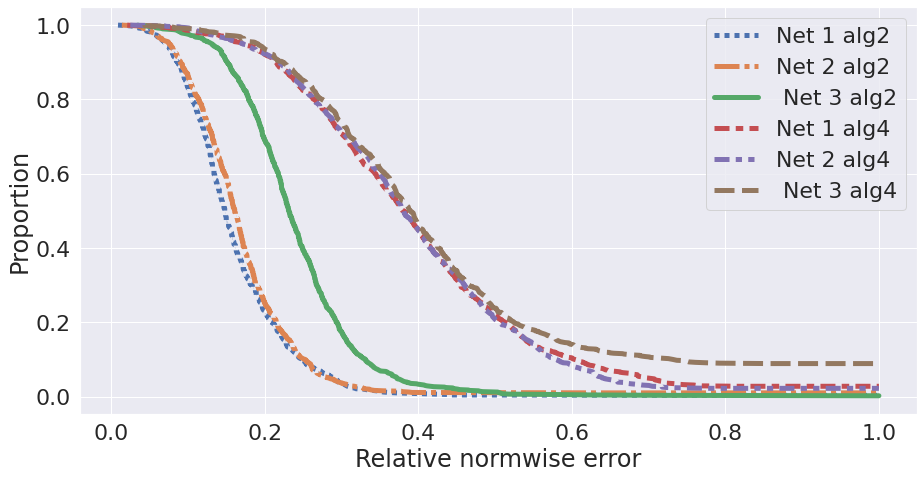}
    \caption{Comparison between the normwise performances of Algorithms 2 and 4 for different networks.
    Axes as for Figure~\ref{fig:plotComp1}.
    }
    \label{fig:plotCompareArchitectures}
\end{figure}

 Figure~\ref{fig:plotCompareArchitectures}
 compares the performance of Algorithms~2 and 4
 on these three networks 
 in attacking 100 images without target, using the
 same measure as Figure~\ref{fig:plotComp1}. 
 We see that for both algorithms changing to a ReLu 
 has little effect.
  Algorithm~2 finds it more difficult to attack Net3 than Net1 or Net2.
  For Algorithm~4, this difference appears only in the tail of the 
  graph; so in moving to a more complex architecture, most images 
  remain just as vulnerable to componentwise attack.

\subsection{Black Box Attacks}\label{subsec:blackbox}
In a black box setting, the attacker does not have access to the inner workings of the network, and hence cannot directly evaluate the Jacobian.
However, with only input and output information it is, of course, possible to approximate the Jacobian using finite differences.
We tested the simple Jacobian approximation 
\begin{equation*}
 \begin{bmatrix}
        \frac{F(x+h\cdot e_1) - F(x)}{h}&\cdots&\frac{F(x+h\cdot e_n) - F(x)}{h}
    \end{bmatrix},
\end{equation*}
where $e_i$ are the standard unit vectors and $h>0$ is a 
small parameter;
we used $h=10^{-3}$.
Measuring the performance of 
Algorithms~2 and 4 on Net1 and Net3 we found that results
with the exact and approximate Jacobian were essentially identical;
see the Supplementary Materials.
We conclude that these attacks work equally well in a black box setting.

\section{Conclusions}\label{sec:conc}
Our aim in this work was to show that it is feasible to  
construct \emph{componentwise} adversarial attacks on image classification systems---here each pixel is perturbed relative to 
a specified tolerance. In particular this allows us to leave 
certain pixels unperturbed.
We developed algorithms  
that build on the normwise approach in \cite{beuzeville:hal-03296180}
and make use of the concept of componentwise backward error from 
\cite{higham_backward_1992}.
Compared with state-of-the-art 
normwise algorithms, when this new approach is applied
to greyscale images with a well-defined background it has the advantage that 
the background can be left unchanged.
In the context of physical writing or printing, such ``adversarial 
ink'' is consistent with a blotchy pen, printer or photocopier.

We illustrated the performance of componentwise attacks on 
three neural networks and in a black box setting.
We also showed that the corresponding 
concept of componentwise condition number has some relevance 
in signalling vulnerability to attack.

Directions for future work include
\begin{itemize}
     \item ~testing the componentwise algorithms on further data sets, notably those involving monochrome images of handwritten or printed text,
     \item ~testing the componentwise algorithms on other image classification tools 
     (note that the algorithms described here do not rely on any 
     specific form for the classification map),
    \item ~the use of object recognition \cite{detect21} to 
    identify background pixels so that the 
    choice of componentwise 
    tolerance vector can be automated in complex images, 
      \item ~the construction of universal 
       componentwise attacks, where the same perturbation 
       changes the classification of many images that have shared
       ``non-background'' locations,
       \item ~the construction of adversarial ink attacks on signatures,
       postcodes, dates, cheques, or entire documents.
\end{itemize}

\section*{Funding}
LB was 
  supported by the MAC-MIGS Centre for Doctoral Training under EPSRC grant EP/S023291/1.
  DJH was supported 
  by EPSRC grants EP/P020720/1 and EP/V046527/1. 
  We thank Oliver Sutton for suggesting the phrase adversarial ink.


\bibliographystyle{siam}
\bibliography{adv_refs}

\begin{thebibliography}{10}

\bibitem{agrawal2018rewriting}
{\sc A.~Agrawal, R.~Verschueren, S.~Diamond, and S.~Boyd}, {\em A rewriting
  system for convex optimization problems}, Journal of Control and Decision, 5
  (2018), pp.~42--60.

\bibitem{Attack_survey_2018}
{\sc N.~Akhtar and A.~Mian}, {\em Threat of adversarial attacks on deep
  learning in computer vision: A survey}, IEEE Access, 6 (2018),
  pp.~14410--14430.

\bibitem{BHV21}
{\sc A.~Bastounis, A.~C. Hansen, and V.~Vla\^ci\'c}, {\em The mathematics of
  adversarial attacks in {AI}--{W}hy deep learning is unstable despite the
  existence of stable neural networks}, arXiv:2109.06098 [cs.LG],  (2021).

\bibitem{beuzeville:hal-03296180}
{\sc T.~Beuzeville, P.~Boudier, A.~Buttari, S.~Gratton, T.~Mary, and
  S.~Pralet}, {\em Adversarial attacks via backward error analysis}.
\newblock hal-03296180, version 3, Dec. 2021.

\bibitem{CorlessNumerical}
{\sc R.~M. Corless and N.~Fillion}, {\em A Graduate Introduction To Numerical
  Methods: From The Viewpoint Of Backward Error Analysis}, Springer, 2013.

\bibitem{diamond2016cvxpy}
{\sc S.~Diamond and S.~Boyd}, {\em {CVXPY}: {A} {P}ython-embedded modeling
  language for convex optimization}, Journal of Machine Learning Research, 17
  (2016), pp.~1--5.

\bibitem{fawzi18}
{\sc A.~Fawzi, O.~Fawzi, and P.~Frossard}, {\em Analysis of classifiers’
  robustness to adversarial perturbations}, Machine Learning, 107 (2018),
  pp.~481--508.

\bibitem{GVl96}
{\sc G.~H. Golub and C.~F. Van~Loan}, {\em Matrix Computations}, The Johns
  Hopkins University Press, fourth~ed., 2013.

\bibitem{deepLearningGoodfellow}
{\sc I.~Goodfellow, Y.~Bengio, and A.~Courville}, {\em Deep learning}, Adaptive
  computation and machine learning, The MIT Press, 2016.

\bibitem{robust}
{\sc I.~J. Goodfellow, P.~D. McDaniel, and N.~Papernot}, {\em Making machine
  learning robust against adversarial inputs}, Commun. {ACM}, 61 (2018),
  pp.~56--66.

\bibitem{harness}
{\sc I.~J. Goodfellow, J.~Shlens, and C.~Szegedy}, {\em Explaining and
  harnessing adversarial examples}, in 3rd International Conference on Learning
  Representations, San Diego, CA, Y.~Bengio and Y.~LeCun, eds., 2015.

\bibitem{HBS09}
{\sc D.~Harrison, T.~M. Burkes, and D.~P. Seiger}, {\em Handwriting
  examination: {M}eeting the challenges of science and the law}, Forensic
  Science Communications, 11 (2009).

\bibitem{higham_backward_1992}
{\sc D.~J. Higham and N.~J. Higham}, {\em Backward error and condition of
  structured linear systems}, SIAM Journal on Matrix Analysis and Applications,
  13 (1992), pp.~162--175.

\bibitem{high:ASNA2}
{\sc N.~J. Higham}, {\em Accuracy and Stability of Numerical Algorithms},
  Society for Industrial and Applied Mathematics, Philadelphia, PA, USA,
  second~ed., 2002.

\bibitem{HH99}
{\sc R.~A. Huber and A.~M. Headrick}, {\em Handwriting Identification: Facts
  and Fundamentals}, CRC Press, Boca Raton, FA, 1999.

\bibitem{lcb-digits}
{\sc Y.~LeCun and C.~Cortes}, {\em {MNIST} handwritten digit database},
  (2010).

\bibitem{mmstv18}
{\sc A.~Madry, A.~Makelov, L.~Schmidt, D.~Tsipras, and A.~Vladu}, {\em Towards
  deep learning models resistant to adversarial attacks}, in 6th International
  Conference on Learning Representations, Vancouver, BC, OpenReview.net, 2018.

\bibitem{M18}
{\sc G.~Marcus}, {\em Deep learning: {A} critical appraisal}, arXiv:1801.00631
  [cs.AI],  (2018).

\bibitem{deepfool}
{\sc S.~Moosavi{-}Dezfooli, A.~Fawzi, and P.~Frossard}, {\em Deepfool: {A}
  simple and accurate method to fool deep neural networks}, in 2016 {IEEE}
  Conference on Computer Vision and Pattern Recognition, NV, USA, {IEEE}
  Computer Society, 2016, pp.~2574--2582.

\bibitem{scsOptimizer}
{\sc B.~O'Donoghue, E.~Chu, N.~Parikh, and S.~Boyd}, {\em Conic optimization
  via operator splitting and homogeneous self-dual embedding}, Journal of
  Optimization Theory and Applications, 169 (2016), pp.~1042--1068.

\bibitem{bb}
{\sc N.~Papernot, P.~D. McDaniel, I.~J. Goodfellow, S.~Jha, Z.~B. Celik, and
  A.~Swami}, {\em Practical black-box attacks against machine learning}, in
  Proceedings of the {ACM} Conference on Computer and Communications Security,
  Abu Dhabi, UAE, R.~Karri, O.~Sinanoglu, A.~Sadeghi, and X.~Yi, eds., {ACM},
  2017, pp.~506--519.

\bibitem{PyTorch19}
{\sc A.~Paszke, S.~Gross, F.~Massa, A.~Lerer, J.~Bradbury, G.~Chanan,
  T.~Killeen, Z.~Lin, N.~Gimelshein, L.~Antiga, A.~Desmaison, A.~Kopf, E.~Yang,
  Z.~DeVito, M.~Raison, A.~Tejani, S.~Chilamkurthy, B.~Steiner, L.~Fang,
  J.~Bai, and S.~Chintala}, {\em {P}y{T}orch: An imperative style,
  high-performance deep learning library}, in Advances in Neural Information
  Processing Systems 32, Curran Associates, Inc., 2019, pp.~8024--8035.

\bibitem{shafahi2018adv}
{\sc A.~Shafahi, W.~Huang, C.~Studer, S.~Feizi, and T.~Goldstein}, {\em Are
  adversarial examples inevitable?}, International Conference on Learning
  Representations, New Orleans, USA,  (2019).

\bibitem{detect21}
{\sc S.~Srivastava, A.~V. Divekar, C.~Anilkumar, I.~Naik, V.~Kulkarni, and
  V.~Pattabiraman}, {\em Comparative analysis of deep learning image detection
  algorithms}, J. Big Data, 8 (2021).

\bibitem{szegedy2013intriguing}
{\sc C.~Szegedy, W.~Zaremba, I.~Sutskever, J.~Bruna, D.~Erhan, I.~Goodfellow,
  and R.~Fergus}, {\em Intriguing properties of neural networks}, arXiv
  preprint arXiv:1312.6199,  (2013).

\bibitem{thwg21}
{\sc I.~Y. Tyukin, D.~J. Higham, A.~Bastounis, E.~Woldegeorgis, and A.~N.
  Gorban}, {\em The feasibility and inevitability of stealth attacks},
  arXiv:2106.13997,  (2021).

\bibitem{tyukin2020adversarial}
{\sc I.~Y. Tyukin, D.~J. Higham, and A.~N. Gorban}, {\em On adversarial
  examples and stealth attacks in artificial intelligence systems}, in 2020
  International Joint Conference on Neural Networks, IEEE, 2020, pp.~1--6.

\bibitem{generalizedInverse}
{\sc G.~Wang, Y.~Wei, and S.~Qiao}, {\em Generalized Inverses: Theory and
  Computations}, Developments in Mathematics, 53, Springer Singapore, 1st ed.
  2018.~ed., 2018.

\end{thebibliography}


\section{Supplementary Material}

\begin{figure}
    \centering
    \begin{subfigure}{\textwidth}
        \centering
        \includegraphics[width=\textwidth]{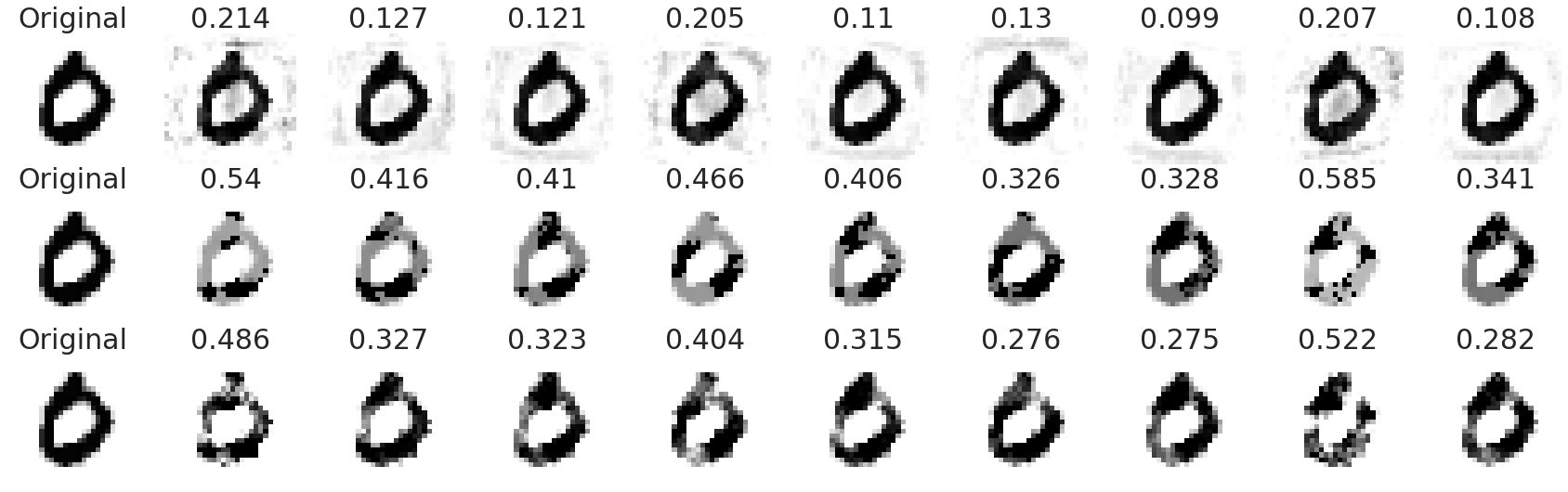}
    \end{subfigure}
    \hfill
    \begin{subfigure}{\textwidth}
        \centering
        \includegraphics[width=\textwidth]{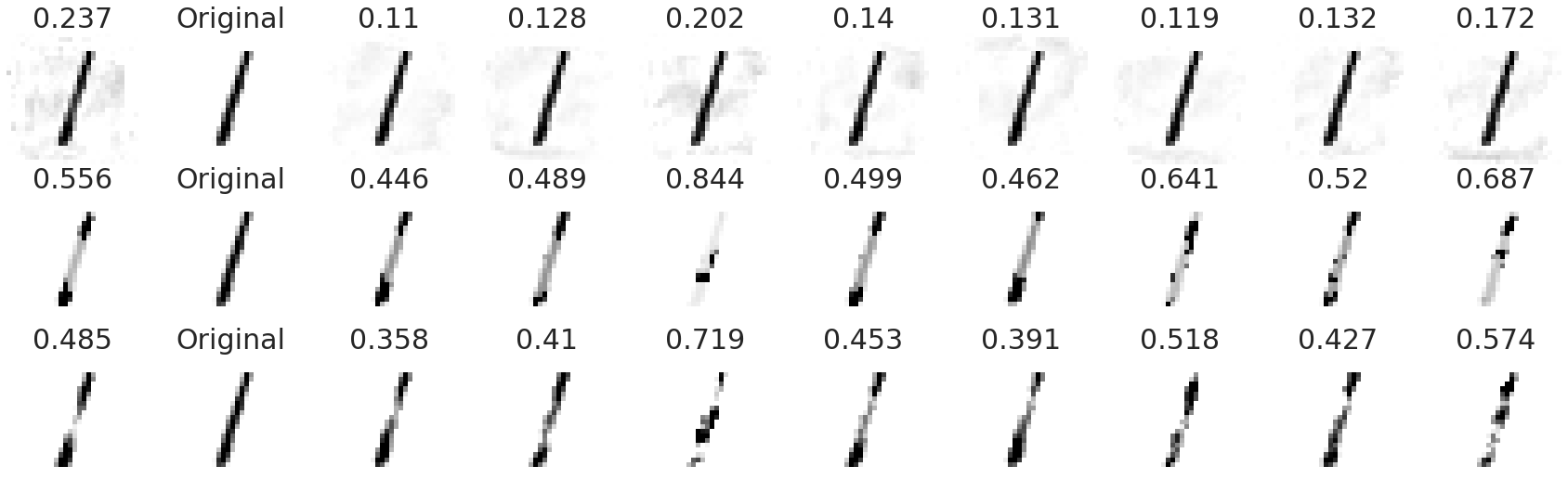}
    \end{subfigure}
    \hfill
    \begin{subfigure}{\textwidth}
        \centering
        \includegraphics[width=\textwidth]{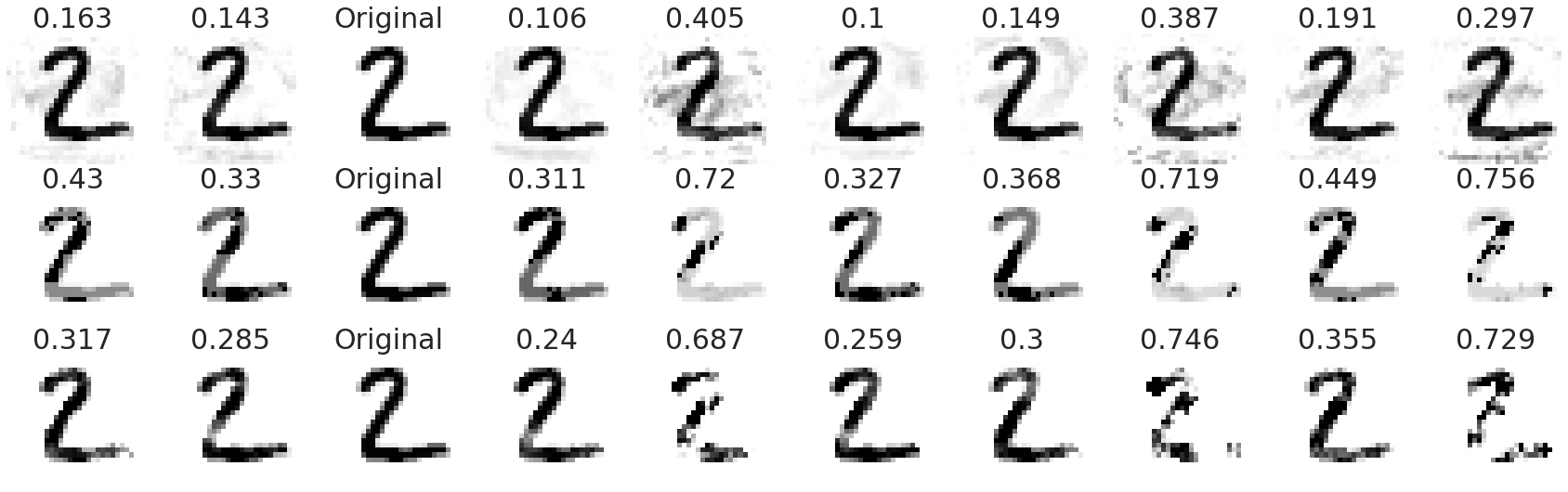}
    \end{subfigure}
    \hfill
    \begin{subfigure}{\textwidth}
        \centering
        \includegraphics[width=\textwidth]{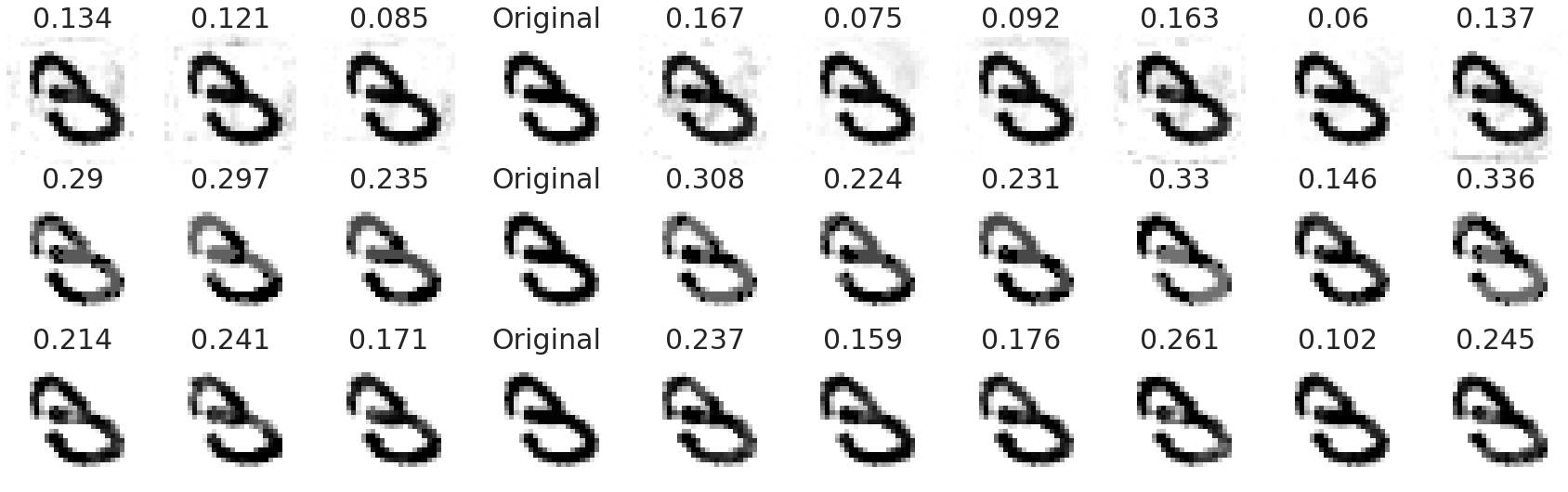}
    \end{subfigure}
    \hfill
    \caption{Row one shows the results of adversarial attacks with 
    Algorithm~2 on an image from class 0. The original image is shown and then, from left to right, we have targets 1,2,3,\ldots,9.
    The numbers above the images indicate the relative 2-norm of the perturbation.
    Rows two and three show this information for  
    Algorithms~3 and 4, respectively.
    This pattern then repeats for images from classes 1,2 and 3.
   }
    \label{fig:img comp 1}
\end{figure}

\begin{figure}
    \centering
    \begin{subfigure}{\textwidth}
        \centering
        \includegraphics[width=\textwidth]{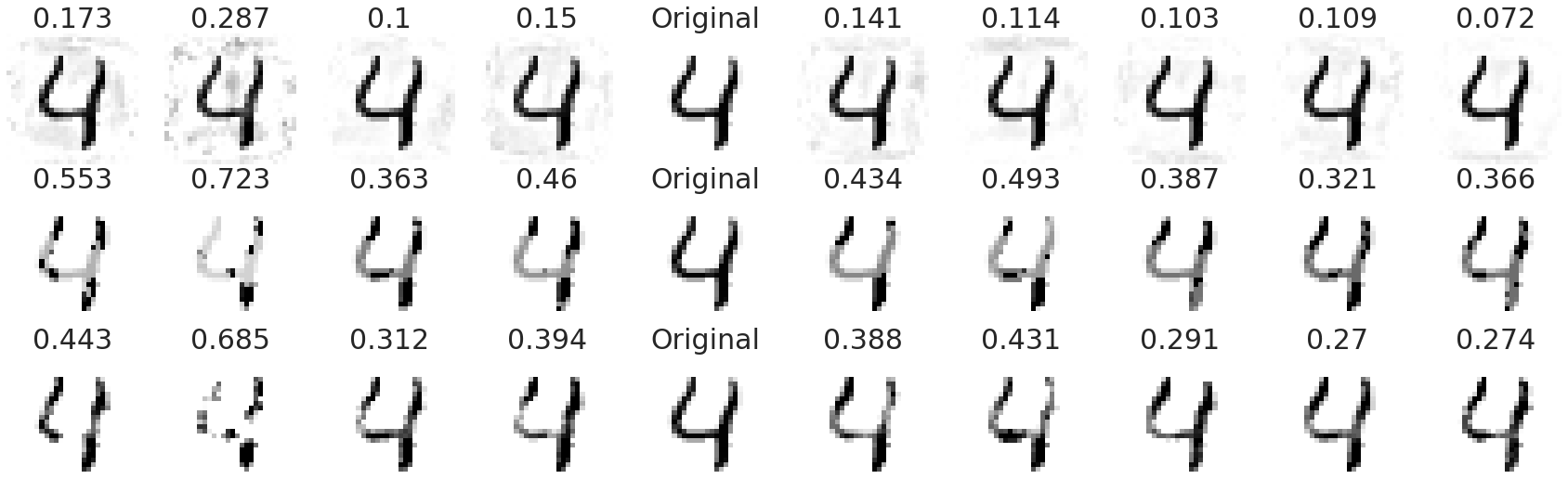}
    \end{subfigure}
    \begin{subfigure}{\textwidth}
        \centering
        \includegraphics[width=\textwidth]{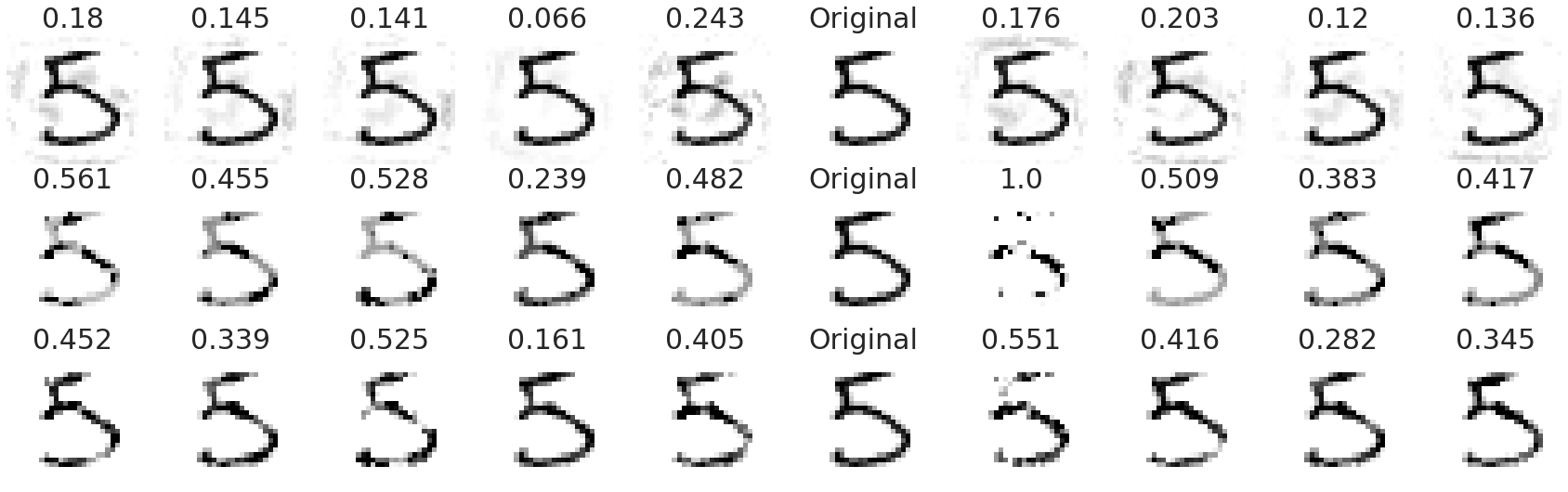}
    \end{subfigure}
    \hfill
    \begin{subfigure}{\textwidth}
        \centering
        \includegraphics[width=\textwidth]{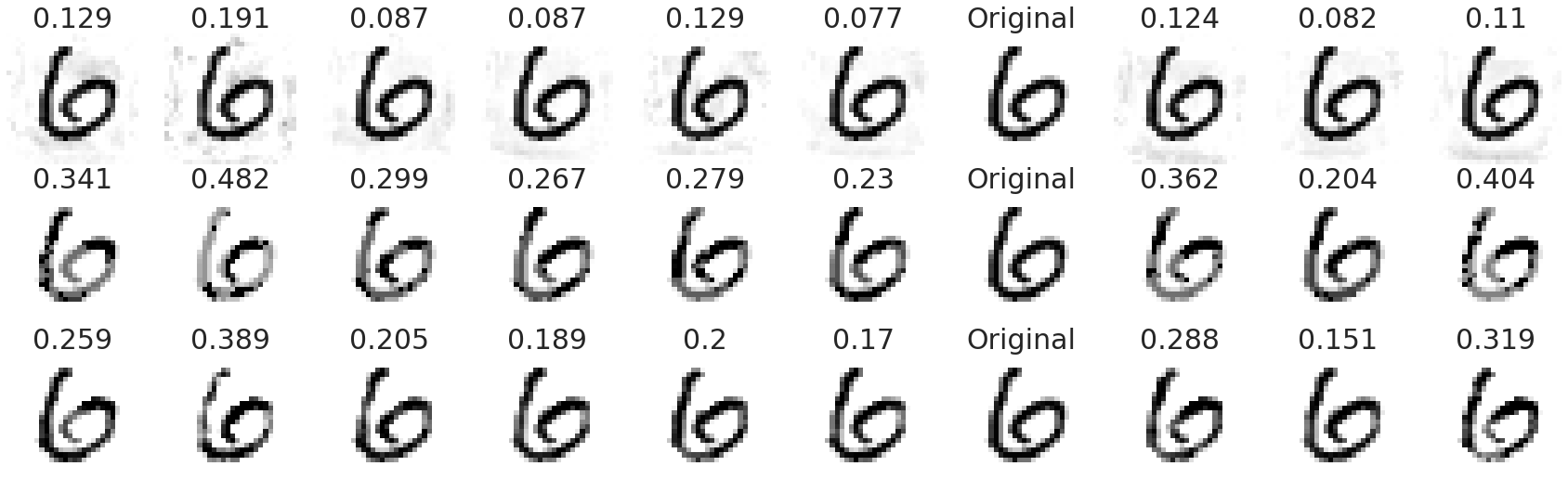}
    \end{subfigure}
    \hfill
    \begin{subfigure}{\textwidth}
        \centering
        \includegraphics[width=\textwidth]{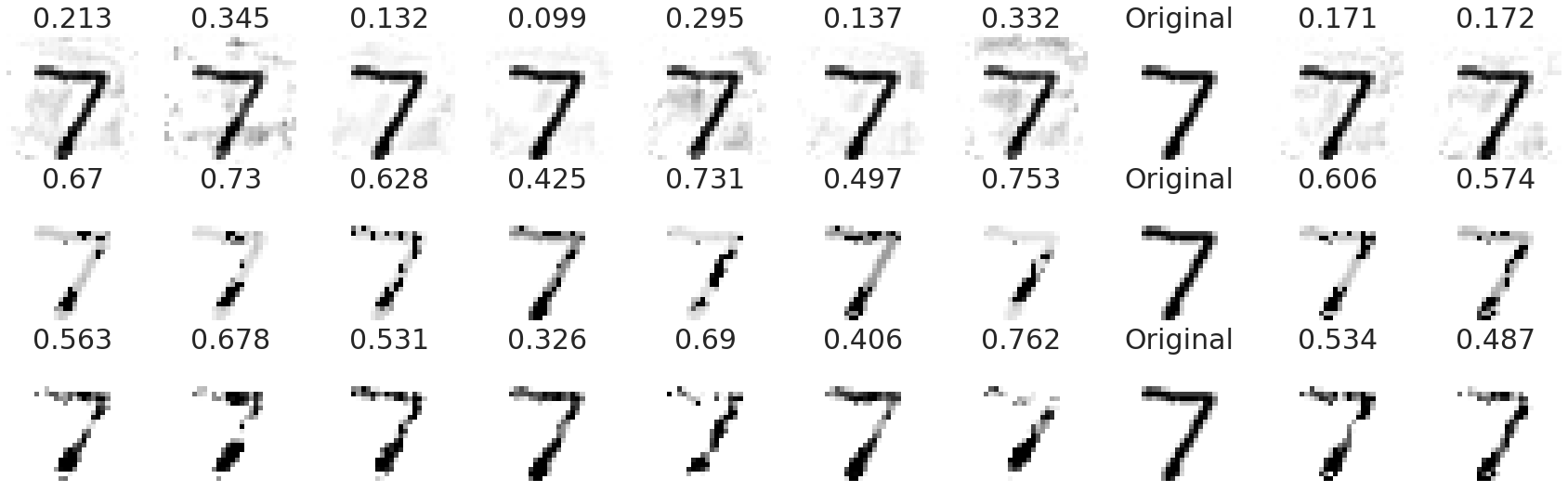}
    \end{subfigure}
    \caption{
    As in Figure~\ref{fig:img comp 1}, rows corresponds to Algorithms~2, 3 and 4 in turn and 
    columns indicate target class. In this case images are from
    classes 4,5,6 and 7.
    }
    \label{fig:img comp 2}
\end{figure}

\begin{figure}
    \centering
    \begin{subfigure}{\textwidth}
        \centering
        \includegraphics[width=\textwidth]{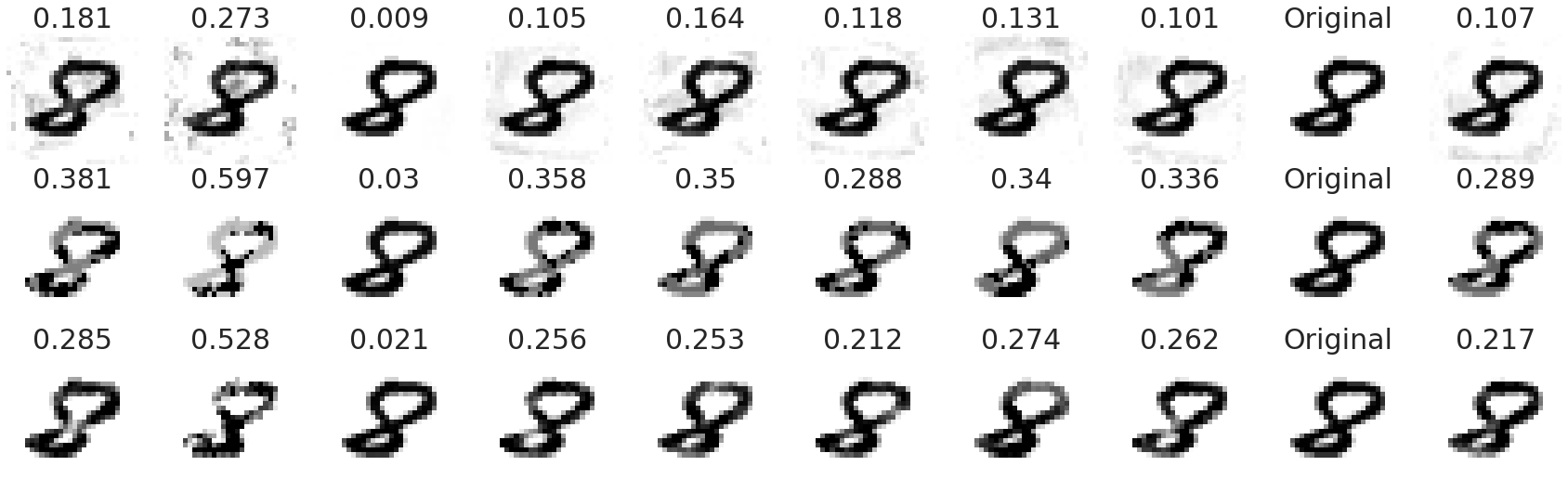}
    \end{subfigure}
    \hfill
    \begin{subfigure}{\textwidth}
        \centering
        \includegraphics[width=\textwidth]{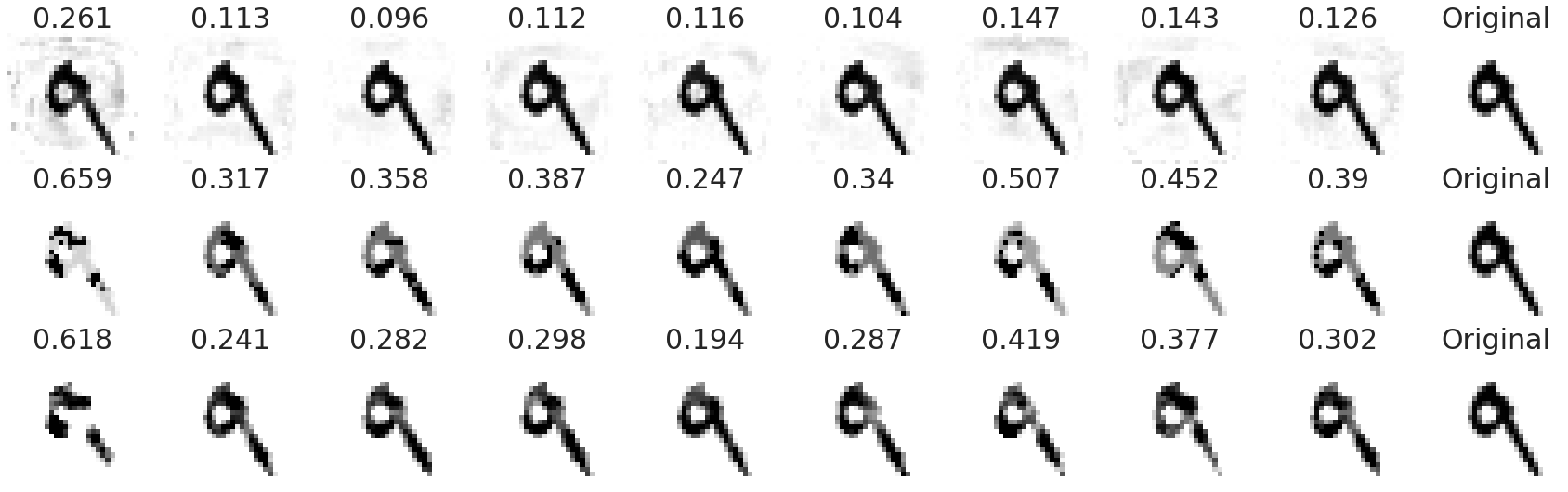}
    \end{subfigure}
    \caption{
     As in Figure~\ref{fig:img comp 1}, rows corresponds to Algorithms~2, 3 and 4 in turn and 
    columns indicate target class. In this case images are from
    classes 8 and 9.
    }
    \label{fig:img comp 3}
\end{figure}

In Figures~\ref{fig:img comp 1},
\ref{fig:img comp 2}
and 
\ref{fig:img comp 3}
we expand
on Figure~\ref{fig:imageComp1.2}
by showing successful attacks 
produced 
by Algorithms~2, 3 and 4, and the relative 2-norm
of the perturbation, on an example from each digit class and for all 
target classes.
Here, each image attacked is the first of its class 
arising in the data set.


Tables~\ref{tab:alg1} and 
\ref{tab:alg3}
are the analogues of 
Tables~\ref{tab:alg2} and \ref{tab:alg4}
corresponding to 
Algorithms~1 and 3, respectively. 

\begin{table}[h]
    \centering
    \begin{tabular}{c|*{10}c}
    & \multicolumn{10}{c}{Alg.\ 1: best target class proportion}\\
    & 0 & 1 & 2 & 3 & 4 & 5 & 6 & 7 & 8 & 9\\ \hline
    0 & 0.00 & 0.00 & 0.17 & 0.05 & 0.00 & \bf{0.38} & 0.14 & 0.11 & 0.00 & 0.15 \\
    1 & 0.02 & 0.00 & 0.04 & \bf{0.4} & 0.00 & 0.08 & 0.06 & 0.29 & 0.08 & 0.02 \\
    2 & 0.03 & 0.03 & 0.00 & \bf{0.51} & 0.00 & 0.02 & 0.09 & 0.09 & 0.22 & 0.01 \\
    3 & 0.02 & 0.02 & 0.22 & 0.00 & 0.00 & \bf{0.54} & 0.01 & 0.04 & 0.13 & 0.03 \\
    4 & 0.03 & 0.00 & 0.05 & 0.01 & 0.00 & 0.00 & 0.04 & 0.15 & 0.09 & \bf{0.64} \\
    5 & 0.02 & 0.00 & 0.02 & \bf{0.49} & 0.02 & 0.00 & 0.13 & 0.13 & 0.15 & 0.03 \\
    6 & 0.04 & 0.01 & 0.20 & 0.01 & 0.02 & \bf{0.52} & 0.00 & 0.07 & 0.01 & 0.12 \\
    7 & 0.01 & 0.02 & \bf{0.41} & 0.33 & 0.02 & 0.00 & 0.00 & 0.00 & 0.01 & 0.21 \\
    8 & 0.02 & 0.01 & 0.32 & \bf{0.37} & 0.01 & 0.09 & 0.02 & 0.08 & 0.00 & 0.07 \\
    9 & 0.00 & 0.01 & 0.02 & 0.10 & 0.23 & 0.10 & 0.00 & \bf{0.41} & 0.14 & 0.00 \\
    \end{tabular}
    \caption{Table of best target proportion for attacks made by Algorithm 1. The rows are the original classes and the columns are the target classes.}
    \label{tab:alg1}
\end{table} 

\begin{table}[h]
    \centering
    \begin{tabular}{c|*{10}c}
    & \multicolumn{10}{c}{Alg.\ 3: best target class proportion}\\
    & 0 & 1 & 2 & 3 & 4 & 5 & 6 & 7 & 8 & 9\\ \hline
    0 & 0.00 & 0.00 & 0.06 & 0.02 & 0.00 & \bf{0.49} & 0.18 & 0.10 & 0.01 & 0.13 \\
    1 & 0.02 & 0.00 & 0.14 & \bf{0.21} & 0.00 & 0.08 & 0.13 & 0.10 & 0.18 & 0.15 \\
    2 & 0.04 & 0.07 & 0.00 & \bf{0.47} & 0.03 & 0.02 & 0.06 & 0.17 & 0.14 & 0.00 \\
    3 & 0.03 & 0.05 & 0.12 & 0.00 & 0.08 & \bf{0.47} & 0.02 & 0.09 & 0.09 & 0.05 \\
    4 & 0.03 & 0.01 & 0.05 & 0.05 & 0.00 & 0.04 & 0.01 & 0.20 & \bf{0.37} & 0.25 \\
    5 & 0.05 & 0.02 & 0.01 & \bf{0.38} & 0.08 & 0.00 & 0.05 & 0.06 & 0.20 & 0.15 \\
    6 & 0.06 & 0.02 & 0.13 & 0.01 & 0.27 & \bf{0.35} & 0.00 & 0.02 & 0.11 & 0.02 \\
    7 & 0.01 & 0.02 & 0.18 & \bf{0.36} & 0.07 & 0.07 & 0.00 & 0.00 & 0.05 & 0.25 \\
    8 & 0.02 & 0.01 & 0.14 & 0.23 & 0.11 & \bf{0.25} & 0.03 & 0.06 & 0.00 & 0.14 \\
    9 & 0.00 & 0.00 & 0.00 & 0.05 & \bf{0.59} & 0.05 & 0.00 & 0.23 & 0.08 & 0.00 \\
    \end{tabular}
    \caption{Table of best target proportion for attacks made by Algorithm 3. The rows are the original classes and the columns are the target classes.}
    \label{tab:alg3}
\end{table}

\begin{figure}
    \centering
    \includegraphics[width=\textwidth]{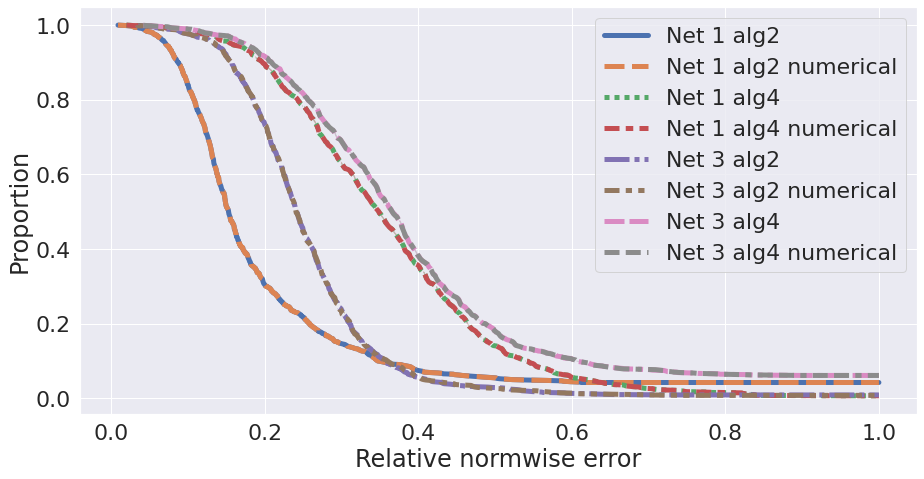}
    \caption{Comparison between the normwise performances of white box and black box attacks for Algorithms~2 and 4 on three different neural networks. 
    Axes as for Figure~\ref{fig:plotComp1}.
    }
    \label{fig:plotNumerical}
\end{figure}

Figure~\ref{fig:plotNumerical} shows the performance measures 
(as described for Figure~\ref{fig:plotComp1}) 
for exact Jacobian and finite-difference (black box)
versions of 
Algorithms~2 and 4 on Net1 and Net3. 

\end{document}